%% file: paper.tex
\documentclass[conference]{IEEEtran}
\IEEEoverridecommandlockouts
\usepackage{cite}
\usepackage{amsmath,amssymb,amsfonts}
\usepackage{algorithmic}
\usepackage{graphicx}
\usepackage{textcomp}
\usepackage{booktabs}
\usepackage{multirow}
\usepackage{float}
\usepackage{xcolor}
\usepackage{url}

\def\BibTeX{{\rm B\kern-.05em{\sc i\kern-.025em b}\kern-.08em
    T\kern-.1667em\lower.7ex\hbox{E}\kern-.125emX}}
\begin{document}

\title{Bridging the Gap: Fusing CNNs and Transformers to Decode the Elegance of Handwritten Arabic Script
}

\author{
    \IEEEauthorblockN{
        Chaouki BOUFENAR,
        Mehdi Ayoub RABIAI, 
        Boualem Nadjib ZAHAF,  
        Khelil Rafik OUARAS
    }  
    \IEEEauthorblockA{\textit{Department of Computer Science} \\
    \textit{University of Algiers 1}\\
    Algiers, Algeria \\
    }
    \IEEEauthorblockA{
        \{c.boufenar, rabiaimehdi, bn.zahaf, kh.ouaras\}@univ-alger.dz  
    }  
}
\maketitle

\begin{abstract}
Handwritten Arabic script recognition is a challenging task due to the script's dynamic letter forms and contextual variations. This paper proposes a hybrid approach combining convolutional neural networks (CNNs) and Transformer-based architectures to address these complexities. We evaluated custom and fine-tuned models, including EfficientNet-B7 and Vision Transformer (ViT-B16), and introduced an ensemble model that leverages confidence-based fusion to integrate their strengths. Our ensemble achieves remarkable performance on the IFN/ENIT dataset, with \textbf{96.38\%} accuracy for letter classification and \textbf{97.22\%}  for positional classification. The results highlight the complementary nature of CNNs and Transformers, demonstrating their combined potential for robust Arabic handwriting recognition. This work advances OCR systems, offering a scalable solution for real-world applications.
\end{abstract}


\section{Introduction}

The recognition of handwritten Arabic script presents a formidable challenge in the field of optical character recognition (OCR). Unlike Latin scripts, Arabic handwriting is characterized by dynamic letter forms, intricate connections, and contextual variations, significantly complicating accurate recognition. Traditional OCR systems, designed primarily for printed text, often falter when confronted with the fluidity and variability inherent in handwritten Arabic. This limitation not only hinders the digitization of historical manuscripts and personal documents but also restricts accessibility for individuals and institutions relying on handwritten records.

In this study, we address these challenges by proposing a novel, intelligent system that leverages the complementary strengths of convolutional neural networks (CNNs) and Transformer-based architectures. CNNs, renowned for their ability to capture fine-grained spatial features, are integrated with Transformers, which excel in modeling global dependencies and contextual relationships. This hybrid approach aims to mirror human perception, enabling the system to analyze both the structural and semantic aspects of Arabic handwriting. By pushing the boundaries of computational intelligence, our goal is to achieve unprecedented accuracy and adaptability in real-world scenarios, thereby bridging the gap between human and machine understanding of handwritten Arabic script.

To ensure robust performance, we utilize a customized dataset based on the IFN/ENIT dataset \cite{Pechwitz2002IFN/ENIT}, a benchmark for Arabic handwriting recognition, which includes a diverse range of handwriting styles and significant intra- and inter-writer variability. Additionally, we employ a comprehensive data augmentation pipeline, incorporating techniques such as elastic deformation, geometric transformations, and Gaussian noise, to enhance model generalization and robustness to real-world variations. These preprocessing steps ensure that our models are well-equipped to handle the complexities of Arabic script, from subtle stroke variations to broader contextual dependencies.

\section{Related Work}

Handwritten Arabic script recognition presents unique challenges due to its cursive nature, contextual variations, and diacritics. Traditional OCR methods relied on handcrafted features and statistical models like Hidden Markov Models (HMMs) and hybrid HMM/ANN systems \cite{8275310, Tagougui2014}, but struggled with the high variability of handwriting.

The rise of deep learning introduced convolutional neural networks (CNNs) as powerful feature extractors, achieving high accuracy on datasets like AHCD and IFN/ENIT \cite{9091836}. However, CNNs face limitations in capturing long-range dependencies, which are crucial for cursive scripts. To address this, Transformer-based models, such as HATFormer \cite{hatformer2024}, leverage self-attention mechanisms for improved recognition. Hybrid approaches combining CNNs with recurrent neural networks (RNNs) \cite{cnn_rnn_ensemble2024} further enhanced performance by integrating local and global features.

Recent research has introduced various deep learning architectures to tackle these challenges, each bringing unique innovations to Arabic handwriting recognition. From Transformer-based models to hybrid CNN-RNN frameworks and novel ensemble methods, different approaches have been proposed to enhance accuracy and robustness. The following works illustrate key advancements in this field.

\subsection{HATFormer: Historic Handwritten Arabic Text Recognition with Transformers}

HATFormer\cite{hatformer2024} propose a Transformer-based encoder-decoder architecture tailored for Arabic handwritten text recognition. Leveraging the Transformer's attention mechanism, HATFormer captures spatial contextual information to address challenges inherent in Arabic script, such as cursive characters and diacritics. Customizations include an image processor for effective ViT preprocessing, a text tokenizer for compact Arabic text representation, and a training pipeline accommodating limited historical Arabic handwriting data.

HATFormer achieved a Character Error Rate (CER) of 8.6\% on the largest public historical handwritten Arabic dataset, marking a 51\% improvement over existing baselines. Additionally, it attained a CER of 4.2\% on a large private non-historical dataset.

\subsection{Arabic Handwriting Recognition System Using Convolutional Neural Network}

This study explores various datasets for Arabic handwriting recognition, including IFN/ENIT and AHDB. The authors review multiple deep learning approaches, emphasizing the effectiveness of CNNs in recognizing Arabic numerals and characters. Notably, a CNN model achieved 94.9\% accuracy on the AHCD database, which contains 16,800 handwritten Arabic characters \cite{arabic_cnn2021}.

The CNN model demonstrated high accuracy in recognizing Arabic characters, highlighting the potential of deep learning techniques in this domain.

\subsection{An End-to-End OCR Framework for Robust Arabic Handwriting Recognition Using a Novel Transformers-Based Model}

Aly Mostafa and al. \cite{ocr_transformer2022} present an end-to-end Optical Character Recognition (OCR) framework employing Vision Transformers (ViT) as an encoder and a vanilla Transformer as a decoder, eliminating the need for CNNs in feature extraction. The study also introduces a 270 million-word multi-font corpus of Classical Arabic with diacritics to enhance training.

The proposed model achieved a CER of 4.46\%, demonstrating the efficacy of Transformer-based architectures in Arabic handwriting recognition.

\subsection{Hybrid Neutrosophic Deep Learning Model for Enhanced Arabic Handwriting Recognition}

This paper introduces a hybrid approach combining Neutrosophic Sets with deep learning models, specifically CNNs integrated with Bidirectional Recurrent Neural Networks (Bi-LSTM and Bi-GRU). This integration aims to enhance the recognition of handwritten Arabic characters by capturing both spatial features and temporal dynamics \cite{neutrosophic2024}.

The NS\_CNN\_Bi-LSTM model achieved an accuracy of 92.38\% on the Hijjaa dataset, while the NS\_CNN\_Bi-GRU model attained 97.38\% accuracy on the AHCD dataset, outperforming previous deep learning approaches.

\subsection{Advancing Arabic Handwriting Recognition with Convolutional and Recurrent Neural Network Ensembles}

The study investigates the application of various deep learning techniques, including CNNs, LSTMs, Bi-LSTMs, GRUs, and Bi-GRUs, to Arabic handwriting recognition. The models were tested on the AHCD and Hijjaa datasets, and their performances were compared using metrics such as Precision, Recall, F1-Score, and Accuracy \cite{cnn_rnn_ensemble2024}.

The Bi-GRU model achieved the highest performance on the AHCD dataset with an accuracy rate of 95.7\%, while the CNN model achieved an accuracy rate of 86.3\% on the Hijjaa dataset.

\section{Overview of the IFN/ENIT Dataset}

The IFN/ENIT dataset \cite{Pechwitz2002IFN/ENIT} is a widely recognized benchmark for handwritten Arabic word recognition, extensively used in Optical Character Recognition (OCR) research. It comprises \textbf{32,489} handwritten Arabic word images stored in BMP format, primarily representing Tunisian city names.

\subsection{Dataset Structure}

The dataset is divided into five subsets (\textbf{set\_a, set\_b, set\_c, set\_d, and set\_e}), each containing unique word samples written by different individuals. This ensures a diverse range of handwriting styles.

\subsection{Variability in Handwriting}

The IFN/ENIT presents significant intra-writer and inter-writer variability, including differences in stroke thickness, letter shapes, spacing, and writing orientation. This variability makes it a challenging benchmark for OCR models, particularly in fine-grained classification tasks such as letter segmentation and position recognition.

\begin{figure}[htbp]
    \centering
    \includegraphics[width=0.3\textwidth]{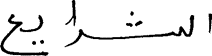}
    \caption{Example of a word image from the IFN/ENIT dataset.}
    \label{fig:ifn_enit}
\end{figure}

\section{Preprocessing Pipeline}

The preprocessing pipeline standardizes handwritten Arabic letter images while preserving character integrity across diverse writing styles. Due to the morphological variability of Arabic script, where letter shapes change based on position, preprocessing ensures consistent input representations without losing positional context.

\subsection{Dataset Restructuring}

To facilitate fine-grained classification, the IFN/ENIT dataset was restructured into an organized corpus of individual letters, preserving their positional context within words. Each sample is categorized based on:

\begin{itemize}
    \item Letter Class: The identity of the Arabic character.
    
    \item Positional Context: Each letter appears in one of four possible forms—
    \textbf{Beginning (B)}, \textbf{Middle (M)}, \textbf{End (E)}, or \textbf{Isolated (I)}.
\end{itemize}

\begin{figure}[htbp]
    \centering
    \includegraphics[width=0.45\textwidth]{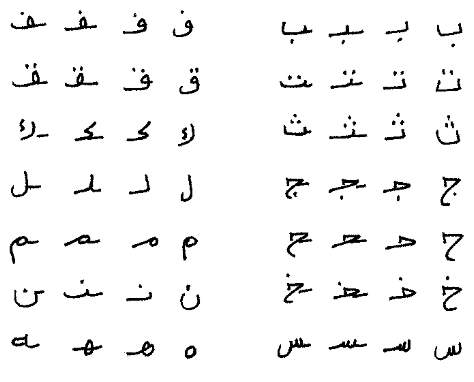}

    \caption{Example of the segmented letters from the IFN/ENIT dataset.}
    \label{fig:ifn_enit_seg}
\end{figure}

\subsection{Image Standardization and Normalization}

Variability in handwriting styles necessitates a structured preprocessing pipeline to ensure consistency while preserving essential structural details:

\begin{itemize}
    \item Grayscale Conversion: All images are converted to grayscale to reduce computational complexity and remove unnecessary color information.
    
    \item Size Standardization: Images are resized to \textbf{128×128 pixels} using a two-step approach:
    
        \begin{itemize}
            \item Padding for Smaller Images: \textbf{White (255)} symmetric padding is applied to preserve the aspect ratio.
            
            \item Resizing for Larger Images: Images exceeding \textbf{128×128} pixels are resized via bilinear interpolation, followed by padding to maintain structural integrity.
        \end{itemize}
    
    \item Normalization: Pixel values are scaled to the \textbf{[0,1]} range by dividing by 255 for consistent intensity distribution.
    
    \item Channel Expansion: To ensure compatibility with deep learning models requiring three-channel inputs, grayscale images are duplicated across three channels while preserving spatial structure.
    
\end{itemize}

\subsection{Composite Labeling}

Each image is annotated with a composite label encoding:

\begin{itemize}
    \item Letter Identity: One of 28 Arabic character classes.
    
    \item Positional Class: One of \{B, M, E, I\}, representing a letter’s position in a word.
\end{itemize}

Instead of treating these as independent classification tasks, the dataset was structured to maintain the inherent correlation between letter identity and positional variation. This approach enables the model to leverage shared structural dependencies, optimizing both classification tasks simultaneously.

\section{Data Augmentation}

To enhance model robustness and improve generalization across diverse handwriting styles, a set of data augmentation techniques was incorporated. These augmentations introduce controlled perturbations that simulate real-world variations in handwritten Arabic characters while preserving structural integrity. The augmentation pipeline consists of the following transformations:

\subsection{Elastic Deformation}

To account for subtle distortions inherent in natural handwriting, elastic deformation was applied:

\begin{itemize}
    \item \textbf{Implementation:} Displacement fields were generated using Gaussian-filtered random noise, which were then applied to the image using bi-directional coordinate mapping.
    \item \textbf{Purpose:} Mimics small-scale variations in stroke curvature and pressure, improving model robustness against intra-class variability.
    \item \textbf{Padding Strategy:} To avoid artifacts at image boundaries, a mirror-padding technique (10 pixels) was applied before transformation, ensuring smooth deformation without the introduction of noise.
\end{itemize}

\subsection{Geometric Transformations}

Handwriting orientation and spatial alignment may vary significantly across different writers. To ensure the model learns invariant representations, random rotations were applied:

\begin{itemize}
    \item \textbf{Rotation Range:} $[-5^\circ, 5^\circ]$.
    \item \textbf{Interpolation Method:} Nearest-neighbor interpolation to maintain character integrity.
    \item \textbf{Fill Value:} A fixed intensity of 255 (white) to match the dataset background.
\end{itemize}

\subsection{Gaussian Blur}

To introduce slight variations in stroke sharpness and simulate optical blurring, Gaussian filtering was employed:

\begin{itemize}
    \item \textbf{Blur Radius:} 0.5 (empirically selected to avoid excessive degradation of fine features).
\end{itemize}

\subsection{Gaussian Noise }

To simulate sensor noise and writing inconsistencies in real-world handwritten text, Gaussian noise was applied:

\begin{itemize}
    \item \textbf{Implementation:} Gaussian noise was added with a mean of 0 and a standard deviation scaled to 40\% of the pixel intensity range ($0.4 \times 255$).
    \item \textbf{Purpose:} Introduces variability in pixel intensities, forcing the model to learn invariant representations robust to environmental distortions.
    \item \textbf{Pixel Clipping:} The noisy image was clipped to maintain valid intensity values within the $[0, 255]$ range.
\end{itemize}

\subsection{Perspective Skew }

To simulate misalignment and irregular handwriting slants, perspective skewing was introduced:

\begin{itemize}
    \item \textbf{Implementation:} A randomized four-point perspective transform was applied while ensuring the character remains centered.
    \item \textbf{Purpose:} Accounts for natural hand tilts, image capture distortions, and irregular paper angles.
    \item \textbf{Re-centering Strategy:} After transformation, a center crop was applied to ensure the character remains within focus.
\end{itemize}

Each augmentation is independently applied to each image with a 30\% probability, meaning a single image may undergo multiple augmentations.

\begin{figure}[htbp]
    \centering
    \includegraphics[width=0.4\textwidth]{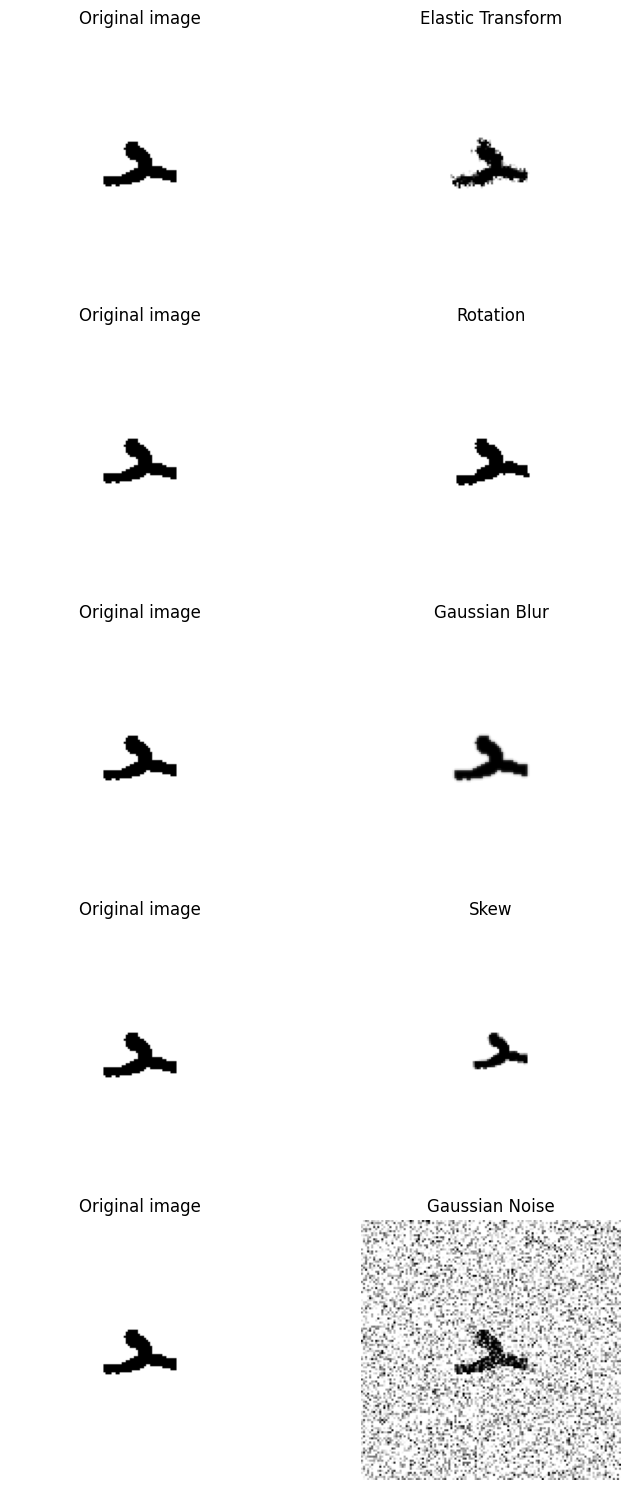}
    \caption{Examples of Each Applied Transformation.}
    \label{fig:Transformation_applie}
\end{figure}

\section{Proposed Approach}

\subsection{Motivation}

Arabic handwriting recognition presents a unique challenge due to the script’s dynamic letter forms and intricate connections. Traditional OCR models struggle to capture these variations, limiting accessibility to handwritten texts.

By integrating CNNs for fine-grained feature extraction and Transformers for global dependencies, we aim to build a model that mirrors human perception—analyzing both structure and meaning. This approach pushes the boundaries of computational intelligence, aiming for higher accuracy and adaptability in real-world scenarios.

\subsection{Architectures}

\subsubsection{Custom CNN}

This model is a custom-built convolutional neural network (CNN) designed for Arabic letter recognition. The architecture is optimized for speed and efficiency while maintaining strong feature extraction capabilities .

\textbf{Architectural Design:}

\begin{itemize}
    \item Feature Extraction: The model employs Two convolutional layers that will extract hierarchical representations of characters.
    
    \item Activation and Pooling: Each convolutional layer is followed by ReLU activations and max-pooling progressively reducing spatial dimensions.
    
    \item Fully Connected Layers: The final feature maps are flattened and passed through two fully connected layers with 128 neurons one layer will produce letters legits and the other will produce position legits which will be transformed to probabilities using the softmax fonction. These fully connected layers were utilized across all model architectures for an optimized multi-task learning framework.
    
    \item Dropout rate of 50\%
    \end{itemize}
\begin{figure}[htbp]
    \centering
    \includegraphics[width=0.5\textwidth]{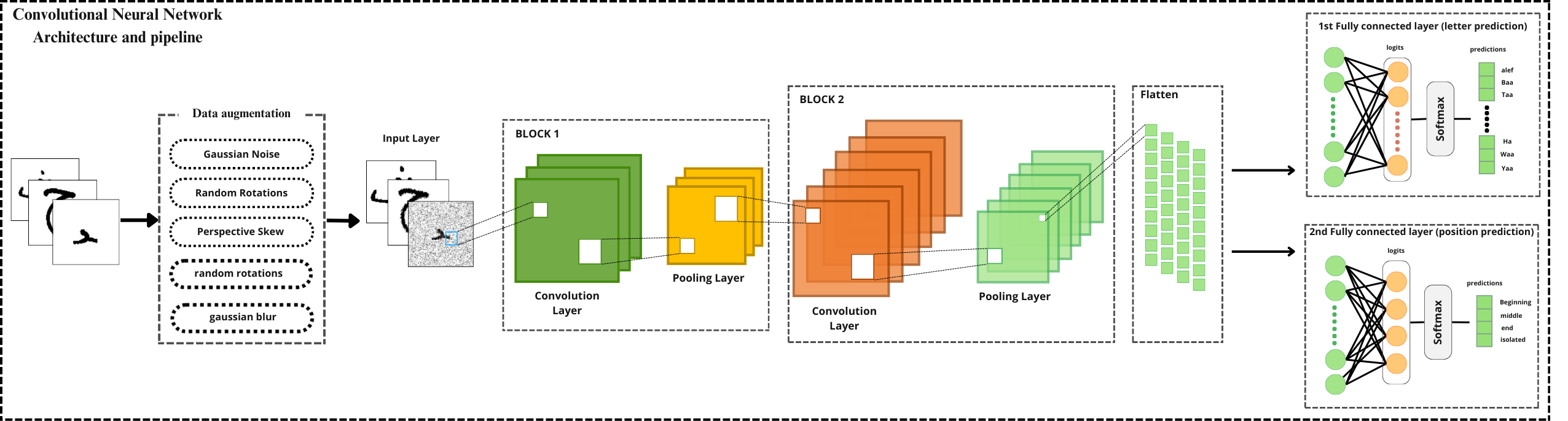}
    \caption{Illustration of the CNN Architecture and pipeline.}
    \label{fig:CNN_architecture}
\end{figure}

\subsubsection{Fine tuning EfficientNet-B7 }

EfficientNet-B7 \cite{Mingxing2019} employs compound scaling principles, balancing depth, width, and resolution, to achieve state-of-the-art performance with fewer parameters compared to conventional architectures. Pre-trained on ImageNet, it serves as a powerful feature extractor, leveraging a hierarchy of MBConv blocks, squeeze-and-excitation mechanisms, and Swish activations to enhance feature representation and capture intricate spatial dependencies. This well structured backbone provides a rich, high-dimensional embedding space that forms the foundation for downstream classification tasks.

\textbf{Architectural Modifications:}

\begin{itemize}
    \item Classifier Removal: Replaced the default classifier with an identity mapping to extract the final embedding, which is then fed into two fully connected layers.  
    \item Dropout rate of 30\%
\end{itemize}
\begin{figure}[htbp]
    \centering
    \includegraphics[width=0.5\textwidth]{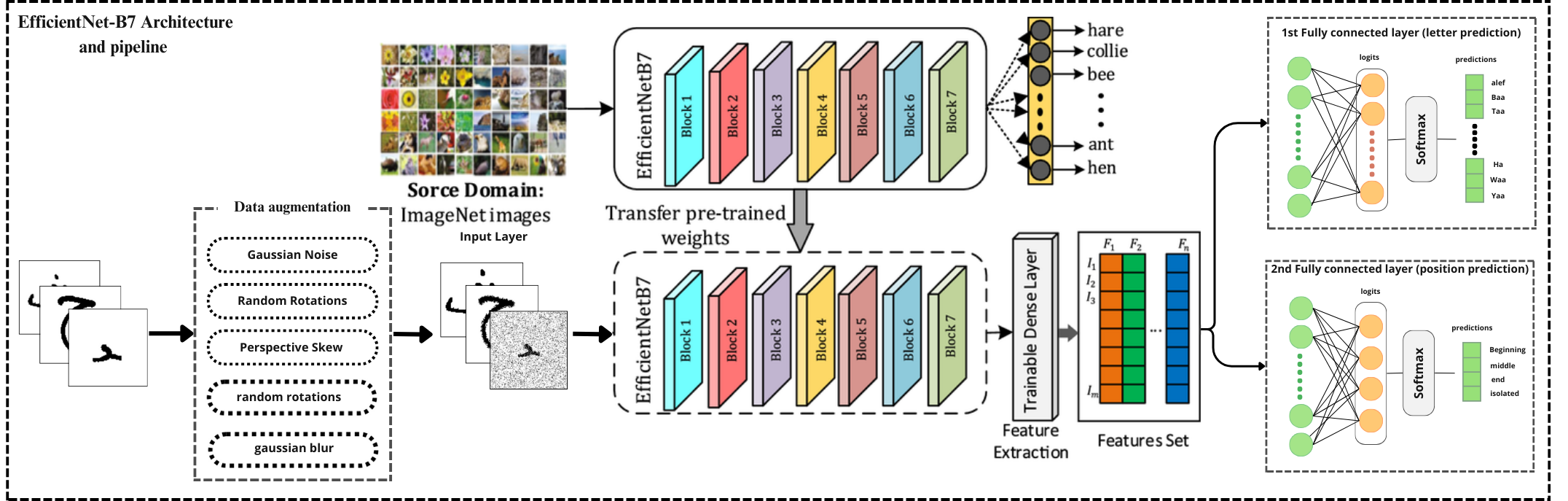}
    \caption{Illustration of the EfficientNet-B7 Architecture and pipeline.}
    \label{fig:EfficientNet_architecture}
\end{figure}

\subsubsection{Custom Vision Transformer }

\begin{figure}[htbp]
    \centering
    \includegraphics[width=0.38\textwidth]{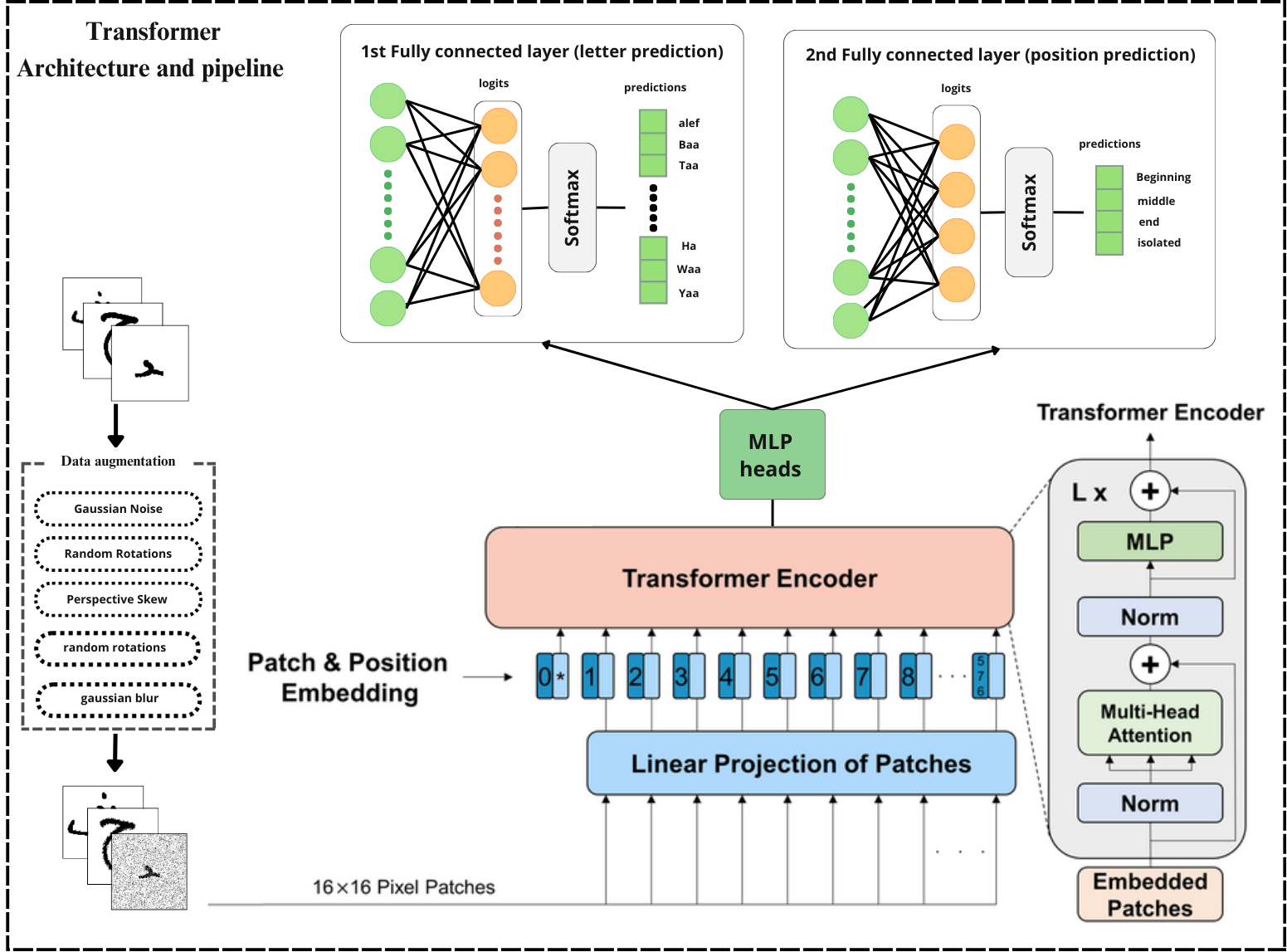}
    \caption{Illustration of the Transformer-Based Architecture and pipeline.}
    \label{fig:Transformer_architecture}
\end{figure}

Unlike CNN-based models, which rely on spatial hierarchies, this Transformer-based architecture learns global dependencies in an image, it tokenizes character images into patches \cite{Alexey2020} and applies self-attention mechanisms, enabling it to model complex variations in Arabic handwriting and efficiently capture long-range dependencies (Fig. \ref{fig:Transformer_architecture}).

\textbf{Architectural Design:}

\begin{itemize}
    \item Patch Embedding: Images are divided into non-overlapping patches, projected into a high-dimensional space, and processed as sequential inputs.
    
    \item Learnable Positional Embeddings: Since the Transformer lacks inherent spatial awareness, a trainable positional encoding matrix is added to the patch embeddings, preserving spatial relationships.
    
    \item Transformer Encoder: 6 layers of multi-head self-attention (MHSA) capture relationships across patches.
    
    \item Layer Normalization and Feedforward Expansion: Each self-attention block is followed by layer normalization and a feedforward network (MLP block).
    
    \item Feature Aggregation: Patch-wise outputs are combined via mean pooling to obtain a single global representation.
    
    \item Multi-Head Self-Attention (MHSA): Each attention head independently captures different aspects of spatial dependencies, enabling the model to recognize subtle variations in Arabic handwriting. This mechanism enhances feature extraction, improves robustness to distortions, and facilitates better contextual understanding across character patches.
    \item Dropout rate of 10\%

\end{itemize}

\vspace{58mm}  

\subsubsection{Fine-Tuned Vision Transformer (ViT-B16)}

The ViT-B16 model follows the same core transformer-based design as our custom architecture but incorporates additional structural and optimization enhancements due to its pretrained nature \cite{Asiri2023}. Instead of learning representations from scratch, ViT-B16 leverages ImageNet-pretrained weights, providing a strong feature initialization.

\textbf{Architectural Modifications:}

\begin{itemize}
    \item Pretrained Feature Extractor: Unlike the custom transformer, which learns all weights during training, ViT-B16 starts with ImageNet-pretrained weights, accelerating convergence and improving generalization.

    \item Positional Embedding Interpolation: The model's original positional encodings were trained for 224×224 inputs. To adapt them to 128×128 images, bicubic interpolation was applied, ensuring minimal distortion of spatial relationships.

    \item Refined Normalization and Scaling: ViT-B16 integrates additional layer normalization steps and a fine-tuned balance of depth and width, optimized through large-scale pretraining.

    \item Classifier Adaptation: The default ViT classification head was removed and replaced with the two independent fully connected layers for letter and position classification.

    \item Dropout rate of 30\%
\end{itemize}

\subsubsection{Ensemble Model Hybrid CNN-Transformer Architecture}

To further refine Arabic letter recognition and mitigate the limitations inherent to individual architectures, an ensemble learning strategy was employed, leveraging the complementary strengths of both convolutional and transformer-based models. Specifically, the fine-tuned EfficientNet-B7 and ViT-B16 models were combined into a unified framework, enhancing both spatial feature extraction and global contextual understanding.

\textbf{Fusion Strategy and Rationale:}

\begin{itemize}
    \item Complementary Strengths: EfficientNet-B7 excels at capturing local structural details through hierarchical feature extraction, whereas the Vision Transformer (ViT-B16) effectively models long-range dependencies and contextual relationships across character strokes.

    \item Confidence-Based Decision Fusion: Rather than treating both models equally, the ensemble dynamically weighs their contributions based on prediction confidence\cite{rosales2023evaluationconfidencebasedensemblingdeep}. Given an input image \(x\), each model produces probability distributions over the possible classes. However, instead of simply averaging these outputs, a confidence score is assigned to each model’s prediction, reflecting how certain it is about its decision.

    The confidence for a given prediction is defined as the highest probability assigned to any class:
    \begin{IEEEeqnarray}{rCl}
        c_{\text{letter}}^{(m)} & = & \max p_{\text{letter}}^{(m)} \quad \\
        c_{\text{position}}^{(m)} & = & \max p_{\text{position}}^{(m)}
    \end{IEEEeqnarray}
    where \( m \in \{E, V\} \) represents either EfficientNet-B7 (\( E \)) or ViT-B16 (\( V \)).

    The final ensemble predictions are obtained by computing a weighted average of each model’s probability distribution:
    \begin{IEEEeqnarray}{rCl}
        p_{\text{letter}}^{\text{ensemble}} & = & \frac{c_{\text{letter}}^{(E)} p_{\text{letter}}^{(E)} + c_{\text{letter}}^{(V)} p_{\text{letter}}^{(V)}}{c_{\text{letter}}^{(E)} + c_{\text{letter}}^{(V)} + \epsilon} \\
        p_{\text{position}}^{\text{ensemble}} & = & \frac{c_{\text{position}}^{(E)} p_{\text{position}}^{(E)} + c_{\text{position}}^{(V)} p_{\text{position}}^{(V)}}{c_{\text{position}}^{(E)} + c_{\text{position}}^{(V)} + \epsilon}
    \end{IEEEeqnarray}
    where \( \epsilon \) is a small constant to prevent division by zero.

    The final predicted class for both letter identity and positional category is then determined as:
    \begin{IEEEeqnarray}{rCl}
        \hat{y}_{\text{letter}} & = & \arg\max p_{\text{letter}}^{\text{ensemble}} \\
        \hat{y}_{\text{position}} & = & \arg\max p_{\text{position}}^{\text{ensemble}}
    \end{IEEEeqnarray}

    \item Regularization Through Model Diversity: By integrating two distinct architectures, the ensemble mitigates biases inherent in each model’s inductive assumptions. thereby reducing systematic misclassifications, particularly in ambiguous handwriting cases.
\end{itemize}

\section{Training Methodology}

To effectively recognize handwritten Arabic letters, the training process is designed to optimize both letter identity and positional classification. This section details the optimization strategy, early stopping criteria, and data-splitting approach used to enhance model performance.

\subsection{\textit{Optimization Strategy}}

To ensure robust learning and generalization, the training pipeline incorporates a structured optimization strategy with early stopping to prevent overfitting. Both CNN and Transformer-based models are trained using multi-task learning, simultaneously optimizing for letter identity and positional classification.

\begin{itemize}
    \item The models are trained using the Adam optimizer with:
    
    \begin{itemize}
        \item Learning rate of \textbf{$10^{-3}$} for the CNN and the EfficientNet-B7 model.
        
        \item Learning rate of \textbf{$10^{-4}$} for the Transformer and the ViT-B16 model.
    \end{itemize}

\item Weighted cross-entropy loss functions are applied separately to the letter and position predictions, with class weights computed from dataset distribution to mitigate class imbalance. The weights are determined based on the inverse class frequency, ensuring that underrepresented classes contribute more significantly to the loss function. The weighting scheme is calculated as follows:

\begin{itemize}
    \item Let \(N\) be the total number of samples in the dataset.
    \item For each class \(c\) in the letter and position classification tasks, the weight \(w_c\) is computed as:

    \[
    w_c = \frac{N}{\text{num\_classes} \times \text{count}(c)}
    \]

    where \(\text{count}(c)\) is the number of occurrences of class \(c\) in the dataset.
    
    \item This ensures that rare classes receive higher weights, preventing the model from biasing toward the more frequent ones.
\end{itemize}

    \item Total loss: Defined as the sum of both objectives, enabling the models to leverage shared representations between the two classification tasks.
\end{itemize}

\subsection{Early Stopping and Model Selection}

\begin{itemize}
    \item Validation loss and accuracy are monitored at each epoch.
    
    \item The best-performing model checkpoint is saved based on the highest \textbf{average accuracy} across letter and position predictions.
    
    \item Training halts if no improvement is observed for \textbf{five consecutive epochs }, preventing unnecessary computations and reducing overfitting risks.(this was only used for the custom models )
\end{itemize}

\subsection{Training Protocol}

\begin{itemize}
    \item The dataset is split into:
    
    \begin{itemize}
        \item \textbf{70\%} training.
        
        \item \textbf{10\%} validation.
        
        \item \textbf{20\%} testing.
    \end{itemize}
    The splitting was performed at the level of pairs, ensuring that each (letter, position) combination was allocated consistently across the training, validation, and test sets.
    \item Batch Size: A batch size of 32 is used for stable gradient updates while maintaining computational efficiency.
\end{itemize}

\section{Results and Discussion}

\subsection{Overall Performance Analysis}

The performance of each model was evaluated based on accuracy, precision, recall, and F1 score for both letter and position classification (Table \ref{tab:performance_metrics}). These results provide insight into the strengths and limitations of individual architectures, as well as the benefits of an ensemble approach.

\begin{table}[htbp]

    \caption{Performance Metrics of Individual Models and Ensemble Strategy for Letter and Position Classification.}
    \centering
    \tiny
    \renewcommand{\arraystretch}{1.3} 
    \setlength{\tabcolsep}{5pt} 
    \begin{tabular}{|c|cc|cc|c|c|}
        \hline
        \textbf{Model} & \multicolumn{2}{c|}{\textbf{Letter Classification }} & \multicolumn{2}{c|}{\textbf{Position Classification}} & \textbf{Test Loss} & \textbf{Training Loss} \\
        \hline
        \multirow{4}{*}{EfficientNet-B7} & Accuracy (\%) & 0.9531 & Accuracy (\%) & 0.9714 & \multirow{4}{*}{0.2352} & \multirow{4}{*}{0.2117} \\
        & Precision (\%) & 0.9548 & Precision (\%) & 0.9715 & & \\
        & Recall (\%) & 0.9531 & Recall (\%) & 0.9714 & & \\
        & F1 Score (\%) & 0.9532 & F1 Score (\%) & 0.9714 & & \\
        \hline
        \multirow{4}{*}{ViT-B/16} & Accuracy (\%) & 0.9431 & Accuracy (\%) & 0.9565 & \multirow{4}{*}{0.3175} & \multirow{4}{*}{0.2409} \\
        & Precision (\%) & 0.9444 & Precision (\%) & 0.9568 & & \\
        & Recall (\%) & 0.9431 & Recall (\%) & 0.9565 & & \\
        & F1 Score (\%) & 0.9432 & F1 Score (\%) & 0.9565 & & \\
        \hline
        \multirow{4}{*}{CNN} & Accuracy (\%) & 0.7891 & Accuracy (\%) & 0.8592 & \multirow{4}{*}{1.0356} & \multirow{4}{*}{1.4923} \\
        & Precision (\%) & 0.7938 & Precision (\%) & 0.8602 & & \\
        & Recall (\%) & 0.7891 & Recall (\%) & 0.8592 & & \\
        & F1 Score (\%) & 0.7895 & F1 Score (\%) & 0.8593 & & \\
        \hline
        \multirow{4}{*}{Transformer Scratch} & Accuracy (\%) & 0.8667 & Accuracy (\%) & 0.9169 & \multirow{4}{*}{0.6258} & \multirow{4}{*}{0.6447} \\
        & Precision (\%) & 0.8699 & Precision (\%) & 0.9171 & & \\
        & Recall (\%) & 0.8667 & Recall (\%) & 0.9169 & & \\
        & F1 Score (\%) & 0.8666 & F1 Score (\%) & 0.9168 & & \\
        \hline
        \multirow{4}{*}{\textbf{Ensemble model}} & \textbf{Accuracy (\%)} & \textbf{0.9638} & \textbf{Accuracy (\%)} & \textbf{0.9722} & \multirow{4}{*}{\textbf{0.1958}} & \multirow{4}{*}{/} \\
        & \textbf{Precision (\%)} & \textbf{0.9643} & \textbf{Precision (\%)} & \textbf{0.9722} & & \\
        & \textbf{Recall (\%)} & \textbf{0.9638} & \textbf{Recall (\%)} & \textbf{0.9722} & & \\
        & \textbf{F1 Score (\%)} & \textbf{0.9638} & \textbf{F1 Score (\%)} & \textbf{0.9722} & & \\
        \hline
    \end{tabular}
    \label{tab:performance_metrics}
\end{table}

\subsubsection{Custom CNN: Performance Trade-offs in Local Feature Learning}

The CNN model reached 78.91\% letter accuracy and 85.92\% position accuracy, demonstrating solid performance in capturing local patterns but facing challenges with more complex variations in handwriting. CNNs inherently prioritize spatial hierarchies, but without additional architectural enhancements or specialized feature extraction techniques, some intricate character details may not be fully captured. The model’s higher The high test loss (1.0356) suggests that while the model effectively learns common structures, it struggles with ambiguous or highly variable letter formations.

\subsubsection{Custom Transformer: Competitive but Impacted by Training Scale}

The scratch-trained transformer model achieved 86.67\% letter accuracy and 91.69\% position accuracy, outperforming the CNN. However, its performance remained below that of both EfficientNet-B7 and ViT-B16, highlighting the challenges of training transformers without extensive pretraining or a large dataset. Transformers typically benefit from vast amounts of data to fully exploit their self-attention mechanisms, and without pretraining, optimization becomes more demanding. The test loss of 0.6258 indicates that while the model captures long-range dependencies, further improvements could be achieved with additional training data or refined initialization strategies.

\subsubsection{EfficientNet-B7: Strong Spatial Feature Extraction}

EfficientNet-B7 demonstrated the highest individual model performance, achieving 95.31\% letter classification accuracy and 97.14\% position classification accuracy. This can be attributed to its hierarchical feature extraction, squeeze-and-excitation mechanisms, and optimized depth-width scaling. The model effectively captures fine-grained spatial details, leading to precise letter recognition and robust positional classification. Its relatively low test loss (0.2352) suggests stable convergence and minimal overfitting.

\subsubsection{ViT-B16: Context-Aware Recognition with Slightly Lower Performance}

The Vision Transformer (ViT-B16) achieved a slightly lower accuracy of 94.31\% for letter classification and 95.65\% for position classification. While it excels at capturing global dependencies, the lack of spatial inductive biases compared to CNN-based models likely contributed to its marginally weaker performance. ViT-B16 also exhibited a higher test loss (0.3175), indicating greater sensitivity to training data variations.

\subsubsection{Ensemble Model: The Best of Both Worlds}

By combining EfficientNet-B7 and ViT-B16 in a confidence-weighted ensemble, we achieved the highest accuracy of 96.38\% for letter classification and 97.22\% for position classification. The ensemble leverages EfficientNet’s superior spatial encoding and ViT-B16’s global contextual reasoning, leading to improved robustness in complex cases. The ensemble also recorded the lowest test loss (0.1958), confirming its ability to generalize more effectively than individual models.

These results highlight the complementary nature of CNN and transformer-based architectures, reinforcing the importance of ensembling to mitigate individual model weaknesses while maximizing their strengths.

\subsection{Per-Letter Accuracy Analysis}

To further analyze model performance, Table \ref{tab:per_letter_accuracy} presents the accuracy of different models for each letter. This comparison highlights variations in recognition accuracy across different Arabic letters.

\begin{table}[htbp]
\scriptsize

\caption{Per-letter accuracy of different models}
\centering
\setlength{\tabcolsep}{3pt} 
\renewcommand{\arraystretch}{1.2} 
\begin{tabular}{|c|c|c|c|c|c|}
    \hline
    \textbf{Letter} & \textbf{CNN} & \textbf{Transformer} & \textbf{EfficientNet-B7} & \textbf{ViT-B/16} & \textbf{Ensemble} \\
    \hline
    Alef  & 0.9404 & 0.9766 & 1.0000 & 0.9936 & 1.0000 \\
    Ayn   & 0.7044 & 0.8289 & 0.9411 & 0.9316 & 0.9563 \\
    Baa   & 0.7886 & 0.7152 & 0.9595 & 0.9772 & 0.9747 \\
    Dad   & 0.7476 & 0.8339 & 0.9425 & 0.9537 & 0.9617 \\
    Dal   & 0.8666 & 0.8680 & 0.9555 & 0.9067 & 0.9484 \\
    Faa   & 0.7340 & 0.8432 & 0.9396 & 0.8955 & 0.9361 \\
    Ghyn  & 0.6730 & 0.7800 & 0.8942 & 0.9049 & 0.9287 \\
    Ha    & 0.6726 & 0.8280 & 0.9677 & 0.9355 & 0.9749 \\
    Haa   & 0.7436 & 0.7474 & 0.9656 & 0.9349 & 0.9554 \\
    Jeem  & 0.8625 & 0.9396 & 0.9955 & 0.9562 & 0.9924 \\
    Kaf   & 0.8245 & 0.9404 & 0.9779 & 0.9830 & 0.9898 \\
    Kha   & 0.7795 & 0.8635 & 0.9383 & 0.9108 & 0.9449 \\
    Lam   & 0.8018 & 0.8671 & 0.9582 & 0.9624 & 0.9574 \\
    Noon  & 0.6756 & 0.8629 & 0.8963 & 0.9448 & 0.9415 \\
    Qaf   & 0.7689 & 0.8042 & 0.9739 & 0.9465 & 0.9791 \\
    Raa   & 0.8217 & 0.9406 & 0.8951 & 0.9021 & 0.9126 \\
    Saad  & 0.7350 & 0.8233 & 0.9767 & 0.9483 & 0.9717 \\
    Seen  & 0.9199 & 0.9124 & 0.9940 & 0.9879 & 0.9970 \\
    Sheen & 0.8978 & 0.9384 & 0.9934 & 0.9882 & 0.9934 \\
    Taa   & 0.7308 & 0.8919 & 0.9195 & 0.9322 & 0.9627 \\
    Thaa  & 0.8300 & 0.8750 & 0.9950 & 0.9583 & 0.9850 \\
    Ttaa  & 0.8768 & 0.9091 & 0.9795 & 0.9824 & 0.9941 \\
    Waw   & 0.8660 & 0.9134 & 0.9918 & 0.9876 & 0.9959 \\
    Yaa   & 0.8083 & 0.9423 & 0.9901 & 0.9684 & 0.9876 \\
    Zay   & 0.8401 & 0.9186 & 0.9622 & 0.9244 & 0.9680 \\
    Dhaa  & 0.8958 & 0.9069 & 0.9845 & 0.9712 & 0.9889 \\
    Dhal  & 0.8733 & 0.8281 & 0.9231 & 0.8846 & 0.9299 \\
    Meem  & 0.7647 & 0.8789 & 0.8733 & 0.8940 & 0.9178 \\
    \hline
\end{tabular}
\label{tab:per_letter_accuracy}
\end{table}

Analyzing per-letter accuracy reveals key performance patterns across models. Structurally simple letters such as \textbf{Alef, Jeem, Seen, and Waw} exhibit consistently high accuracy across all architectures, with the ensemble model achieving near-perfect classification. These characters have distinct shapes with minimal variation, making them easier to recognize.

The \textbf{CNN model}, constrained by local feature extraction, struggles with letters that exhibit high intra-class variability, such as \textbf{Ayn, Dad, Ghyn, and Noon}. These characters undergo substantial shape transformations depending on their position within a word, making them difficult to generalize using purely spatial convolutional filters. The lack of an explicit mechanism for capturing long-distance dependencies further exacerbates misclassification in cases where contextual relationships are crucial.

Introducing \textbf{self-attention mechanisms}, the \textbf{custom Transformer model} mitigates some of these issues, demonstrating improved accuracy across several ambiguous letters. However, training a transformer from scratch without large-scale dataset and  pretraining introduces instability, leading to moderate performance drops on complex shapes where global context alone is insufficient.

\textbf{EfficientNet-B7}, leveraging pre-trained hierarchical feature extraction, significantly improves recognition of fine details in challenging letters. However, it remains inconsistent when classifying letters with significant morphological variations across different positional contexts, where broader contextual awareness is essential for accurate identification.

\textbf{ViT-B16}, pre-trained on large-scale datasets, achieves a strong balance between local and global features, surpassing both the CNN and Transformer models. While pretraining helps balance global and local representations, reliance on positional embeddings limits ViT’s ability to differentiate fine-grained stroke-level variations, particularly in handwritten forms.

Ultimately, the \textbf{ensemble model} consistently achieves the highest per-letter accuracy by balancing both local feature extraction and global attention mechanisms. It successfully mitigates the individual weaknesses of each architecture, demonstrating superior performance on ambiguous letters like \textbf{Dad, Ghyn and Noon}, where neither local nor global features alone were sufficient.

\subsection{Per-Pair Accuracies}

Beyond individual letter classification, we evaluate model performance in correctly predicting both the letter and its positional context. as shown in Table \ref{tab:accuracies_cleaned}. This metric highlights how well each model adapts to the morphological variations Arabic letters undergo across different positions within a word.

\begin{table}[htbp]
\centering

\caption{Accuracies for pairwise Letter and Position Predictions.}
\tiny 
\begin{tabular}{|c|c|c|c|c|c|c|}
\hline
\multirow{2}{*}{Letter} & \multirow{2}{*}{Position} & \multicolumn{5}{c|}{Models} \\
\cline{3-7}
 & & CNN & EFICNETNETB7 & VITB16 & TRANSFORMER & ensemble modele  \\
\hline
\multirow{2}{*}{Alef} 
    & E & 82.80\% & 97.45\% & 96.18\% & 92.99\% & 98.09\% \\
    & I & 94.25\% & 99.68\% & 98.72\% & 97.12\% & 99.68\% \\
\hline
\multirow{4}{*}{Ayn} 
    & B & 43.55\% & 97.21\% & 95.12\% & 78.05\% & 97.21\% \\
    & E & 58.28\% & 93.87\% & 94.48\% & 76.69\% & 96.93\% \\
    & I & 69.97\% & 94.54\% & 95.90\% & 82.94\% & 96.59\% \\
    & M & 69.90\% & 88.03\% & 88.03\% & 71.52\% & 93.20\% \\
\hline
\multirow{4}{*}{Baa} 
    & B & 58.97\% & 91.28\% & 92.31\% & 52.31\% & 94.36\% \\
    & E & 67.09\% & 87.97\% & 93.04\% & 65.19\% & 90.51\% \\
    & I & 79.33\% & 94.67\% & 94.00\% & 88.00\% & 96.67\% \\
    & M & 71.43\% & 95.12\% & 95.82\% & 69.69\% & 96.17\% \\
\hline
\multirow{4}{*}{Dad} 
    & B & 56.29\% & 95.36\% & 91.39\% & 78.81\% & 94.70\% \\
    & E & 93.06\% & 98.84\% & 97.69\% & 91.91\% & 98.84\% \\
    & I & 80.67\% & 92.00\% & 98.00\% & 85.33\% & 98.00\% \\
    & M & 44.74\% & 86.84\% & 87.50\% & 61.18\% & 89.47\% \\
\hline
\multirow{2}{*}{Dal} 
    & I & 80.55\% & 94.76\% & 85.54\% & 90.52\% & 94.02\% \\
    & E & 77.70\% & 92.91\% & 83.78\% & 75.68\% & 91.22\% \\
\hline
\multirow{4}{*}{Faa} 
    & B & 62.94\% & 88.83\% & 83.25\% & 73.60\% & 90.36\% \\
    & E & 55.30\% & 86.36\% & 85.61\% & 68.94\% & 87.12\% \\
    & I & 70.67\% & 93.33\% & 95.33\% & 86.00\% & 96.67\% \\
    & M & 76.70\% & 94.24\% & 86.13\% & 83.25\% & 92.15\% \\
\hline
\multirow{4}{*}{Ghyn} 
    & B & 30.81\% & 97.67\% & 88.95\% & 73.26\% & 95.93\% \\
    & E & 37.24\% & 82.76\% & 82.76\% & 66.21\% & 85.52\% \\
    & I & 73.79\% & 97.41\% & 96.44\% & 84.79\% & 97.41\% \\
    & M & 60.93\% & 74.42\% & 88.37\% & 65.58\% & 84.65\% \\
\hline
\multirow{4}{*}{HA} 
    & B & 69.67\% & 95.08\% & 92.08\% & 75.96\% & 95.08\% \\
    & E & 34.00\% & 96.67\% & 95.33\% & 79.33\% & 97.33\% \\
    & I & 68.67\% & 96.00\% & 95.33\% & 92.00\% & 96.67\% \\
    & M & 56.73\% & 93.57\% & 92.98\% & 72.51\% & 95.32\% \\
\hline
\multirow{4}{*}{Haa} 
    & B & 63.59\% & 92.39\% & 89.67\% & 75.00\% & 95.11\% \\
    & E & 58.67\% & 95.33\% & 90.00\% & 64.00\% & 96.00\% \\
    & I & 76.00\% & 98.00\% & 97.33\% & 74.67\% & 98.67\% \\
    & M & 53.33\% & 92.67\% & 88.67\% & 78.00\% & 90.67\% \\
\hline
\multirow{4}{*}{Jeem} 
    & B & 63.33\% & 97.33\% & 90.67\% & 78.67\% & 97.33\% \\
    & E & 86.00\% & 97.33\% & 94.00\% & 93.33\% & 97.33\% \\
    & I & 66.67\% & 97.33\% & 94.00\% & 90.67\% & 98.67\% \\
    & M & 74.53\% & 99.06\% & 95.75\% & 91.98\% & 99.06\% \\
\hline
\multirow{4}{*}{Kaf} 
    & B & 75.32\% & 95.45\% & 97.40\% & 89.61\% & 96.75\% \\
    & E & 60.66\% & 96.72\% & 96.72\% & 86.07\% & 98.36\% \\
    & I & 84.67\% & 99.33\% & 99.33\% & 96.00\% & 99.33\% \\
    & M & 65.22\% & 99.38\% & 98.14\% & 89.44\% & 99.38\% \\
\hline
\multirow{4}{*}{Kha} 
    & B & 55.28\% & 83.23\% & 79.50\% & 71.43\% & 88.20\% \\
    & E & 67.33\% & 94.00\% & 86.67\% & 69.33\% & 92.00\% \\
    & I & 78.00\% & 97.67\% & 95.00\% & 87.67\% & 97.33\% \\
    & M & 54.30\% & 91.39\% & 86.09\% & 76.82\% & 92.72\% \\
\hline
\multirow{4}{*}{Lam} 
    & B & 66.99\% & 86.06\% & 92.91\% & 74.33\% & 92.18\% \\
    & E & 72.96\% & 95.41\% & 94.90\% & 81.63\% & 93.37\% \\
    & I & 93.00\% & 99.33\% & 96.67\% & 93.33\% & 99.00\% \\
    & M & 65.29\% & 95.88\% & 96.91\% & 87.63\% & 97.59\% \\
\hline
\multirow{4}{*}{Noon} 
    & B & 71.81\% & 89.26\% & 94.63\% & 80.54\% & 95.30\% \\
    & E & 70.67\% & 92.00\% & 96.67\% & 89.33\% & 96.67\% \\
    & I & 78.67\% & 94.67\% & 96.67\% & 88.00\% & 98.00\% \\
    & M & 41.61\% & 73.83\% & 84.56\% & 77.85\% & 83.22\% \\
\hline
\multirow{4}{*}{QAF} 
    & B & 81.82\% & 95.29\% & 94.61\% & 78.45\% & 96.30\% \\
    & E & 80.00\% & 91.33\% & 92.00\% & 62.67\% & 92.00\% \\
    & I & 58.00\% & 98.67\% & 90.67\% & 86.00\% & 98.67\% \\
    & M & 61.54\% & 92.90\% & 89.94\% & 76.33\% & 94.08\% \\
\hline
\multirow{2}{*}{Raa} 
    & E & 89.33\% & 97.33\% & 98.00\% & 96.00\% & 97.33\% \\
    & I & 47.79\% & 78.68\% & 63.97\% & 86.76\% & 74.26\% \\
\hline
\multirow{4}{*}{Saad} 
    & B & 61.33\% & 92.00\% & 88.67\% & 72.00\% & 92.00\% \\
    & E & 82.67\% & 98.67\% & 96.67\% & 78.00\% & 98.00\% \\
    & I & 76.00\% & 99.33\% & 98.00\% & 86.00\% & 100.00\% \\
    & M & 48.67\% & 90.00\% & 87.33\% & 70.00\% & 92.00\% \\
\hline
\multirow{4}{*}{Seen} 
    & B & 88.67\% & 99.01\% & 98.52\% & 90.15\% & 99.01\% \\
    & E & 94.34\% & 97.48\% & 95.60\% & 82.39\% & 97.48\% \\
    & I & 68.00\% & 99.33\% & 96.00\% & 82.00\% & 97.33\% \\
    & M & 92.00\% & 98.67\% & 97.33\% & 83.33\% & 98.00\% \\
\hline
\multirow{4}{*}{Sheen} 
    & B & 79.01\% & 96.95\% & 97.33\% & 94.27\% & 98.09\% \\
    & E & 84.77\% & 96.69\% & 94.04\% & 91.39\% & 97.35\% \\
    & I & 76.00\% & 96.00\% & 98.67\% & 88.00\% & 98.00\% \\
    & M & 85.50\% & 96.50\% & 96.50\% & 80.50\% & 97.00\% \\
\hline
\multirow{4}{*}{Taa} 
    & B & 79.24\% & 94.61\% & 94.61\% & 92.02\% & 97.21\% \\
    & E & 38.41\% & 79.47\% & 82.12\% & 63.58\% & 84.77\% \\
    & I & 58.00\% & 89.33\% & 96.00\% & 89.33\% & 96.00\% \\
    & M & 65.74\% & 84.26\% & 87.50\% & 85.19\% & 91.20\% \\
\hline
\multirow{4}{*}{Thaa} 
    & B & 63.33\% & 98.00\% & 94.67\% & 71.33\% & 100.00\% \\
    & E & 87.33\% & 93.33\% & 96.00\% & 81.33\% & 97.33\% \\
    & I & 74.67\% & 98.00\% & 91.33\% & 77.33\% & 96.67\% \\
    & M & 75.33\% & 97.33\% & 88.00\% & 82.00\% & 94.00\% \\
\hline
\multirow{3}{*}{Ttaa} 
    & E & 55.32\% & 100.00\% & 100.00\% & 72.34\% & 97.87\% \\
    & I & 56.00\% & 98.67\% & 96.67\% & 84.00\% & 100.00\% \\
    & M & 71.53\% & 90.28\% & 85.42\% & 79.86\% & 90.28\% \\
\hline
\multirow{2}{*}{Waw} 
    & E & 74.05\% & 96.76\% & 97.84\% & 76.76\% & 98.92\% \\
    & I & 78.33\% & 98.00\% & 93.33\% & 94.33\% & 97.33\% \\
\hline
\multirow{4}{*}{Yaa} 
    & B & 66.47\% & 97.01\% & 95.61\% & 83.43\% & 97.21\% \\
    & E & 51.46\% & 79.08\% & 80.33\% & 50.63\% & 83.68\% \\
    & I & 67.31\% & 92.58\% & 74.73\% & 86.54\% & 88.19\% \\
    & M & 76.97\% & 98.43\% & 92.13\% & 90.75\% & 97.24\% \\
\hline
\multirow{2}{*}{Zay} 
    & E & 68.56\% & 94.85\% & 85.05\% & 81.96\% & 93.30\% \\
    & I & 73.33\% & 97.33\% & 92.67\% & 91.33\% & 95.33\% \\
\hline
\multirow{3}{*}{dhaa} 
    & B & 78.00\% & 84.67\% & 86.67\% & 70.67\% & 90.00\% \\
    & I & 56.67\% & 96.00\% & 94.00\% & 91.33\% & 95.33\% \\
    & M & 85.43\% & 92.05\% & 87.42\% & 78.15\% & 93.38\% \\
\hline
\multirow{2}{*}{dhal} 
    & I & 58.23\% & 94.38\% & 85.54\% & 81.12\% & 90.36\% \\
    & E & 67.88\% & 88.08\% & 74.61\% & 73.58\% & 85.49\% \\
\hline
\multirow{4}{*}{meem} 
    & B & 66.37\% & 83.27\% & 85.79\% & 83.63\% & 88.67\% \\
    & E & 57.79\% & 88.96\% & 88.96\% & 72.73\% & 90.91\% \\
    & I & 75.84\% & 97.32\% & 96.64\% & 81.21\% & 97.99\% \\
    & M & 75.24\% & 85.71\% & 87.48\% & 79.18\% & 88.98\% \\
\hline
\end{tabular}
\label{tab:accuracies_cleaned}
\end{table}

For CNN-based architectures (EfficientNet and the custom CNN), pairwise accuracy remained notably low for letters exhibiting substantial shape transformations between positions—an expected limitation of CNNs, which rely on local feature extraction. However, EfficientNet’s pretraining significantly improved recognition, even for highly variable letters. For instance, the letter (Ghyn) in the \textbf{beginning} and \textbf{end}  positions showed a remarkable improvement from \textbf{30.81\%} and \textbf{37.24\%} in the custom CNN to \textbf{97.67\%} and \textbf{82.76\%} in EfficientNet.

While EfficientNet achieved over \textbf{90\%} accuracy for most letter pairs, it struggled with letters that undergo radical transformations, such as (Ghyn) and (Noon) in their \textbf{medial}  position, where accuracy dropped to \textbf{74.42\%} and \textbf{73.83\%}, respectively. This is due to the significant structural differences these letters exhibit across positions, a challenge for CNNs that primarily capture spatial hierarchies without strong contextual awareness.

This is where Transformer-based architectures excel. Their attention mechanisms allow them to handle extreme variations more effectively than CNNs. The \textbf{custom Transformer} model demonstrated substantial gains for high-variability letters compared to the CNN model. For example, (Ghyn) in \textbf{beginning}  and \textbf{end} positions improved from \textbf{30.81\%} and \textbf{37.24\%} (CNN) to \textbf{73.26\%} and \textbf{66.21\%} (Transformer). Similarly, \textbf{ViT-B16} outperformed EfficientNet on highly variable classes, further reinforcing the strength of global attention in learning structural transformations.

However, despite their advantage in capturing global dependencies, Transformers exhibited weaknesses in fine-grained stroke differentiation, particularly in cases where small ligatures determine positional identity. For instance, (Yaa) in the \textbf{isolated}  position was classified with \textbf{74.73\%} accuracy in \textbf{ViT-B16}, whereas EfficientNet achieved \textbf{92.58\%} due to CNNs’ ability to capture fine local details which was particularly beneficial for the letter \textbf{Yaa}, as the difference between its \textbf{isolated} and \textbf{final} forms is minimal, with the only distinguishing feature being the connecting stroke that links it to the preceding letter in the final position.

By combining the strengths of both architectures, the \textbf{ensemble model} achieved a well-balanced performance, mitigating weaknesses in both local and global feature extraction. This resulted in a \textbf{minimum per-pair accuracy of 74.26\%}, with the majority exceeding \textbf{90\%}, demonstrating the ensemble’s ability to generalize across both fine-grained and large-scale variations in letter morphology.

\begin{figure}[htbp]
    \centering
    \includegraphics[width=0.5\textwidth]{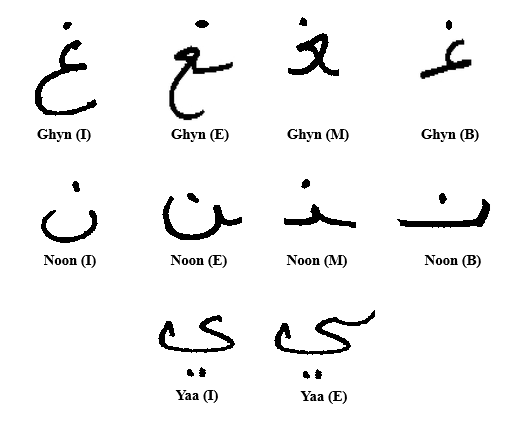}

    \caption{Examples of morphological variations in letters due to positional changes.}
    \label{fig:Example_L_P}
\end{figure}
\section{Conclusion}
This study presents a hybrid approach for handwritten Arabic script recognition, integrating CNNs and Transformer-based architectures to address the script's dynamic letter forms and contextual variations. The proposed ensemble model achieves high performance, with 96.38\%  accuracy for letter classification and 97.22\%  accuracy for positional classification on the IFN/ENIT dataset. These results demonstrate the complementary strengths of CNNs in spatial feature extraction and Transformers in capturing global contextual information.

The ensemble model, which employs Confidence-Based Fusion of EfficientNet-B7 and ViT-B16, exhibits robustness in recognizing Arabic letters with significant intra-class variability, such as Ghyn and Noon. By effectively balancing local and global feature extraction, the model generalizes well across diverse handwriting styles and positional variations. However, certain misclassifications occur, not necessarily due to algorithmic limitations but rather the inherent complexity of handwritten Arabic script.

\begin{figure}[htbp]
    \centering
    \includegraphics[width=0.49\textwidth]{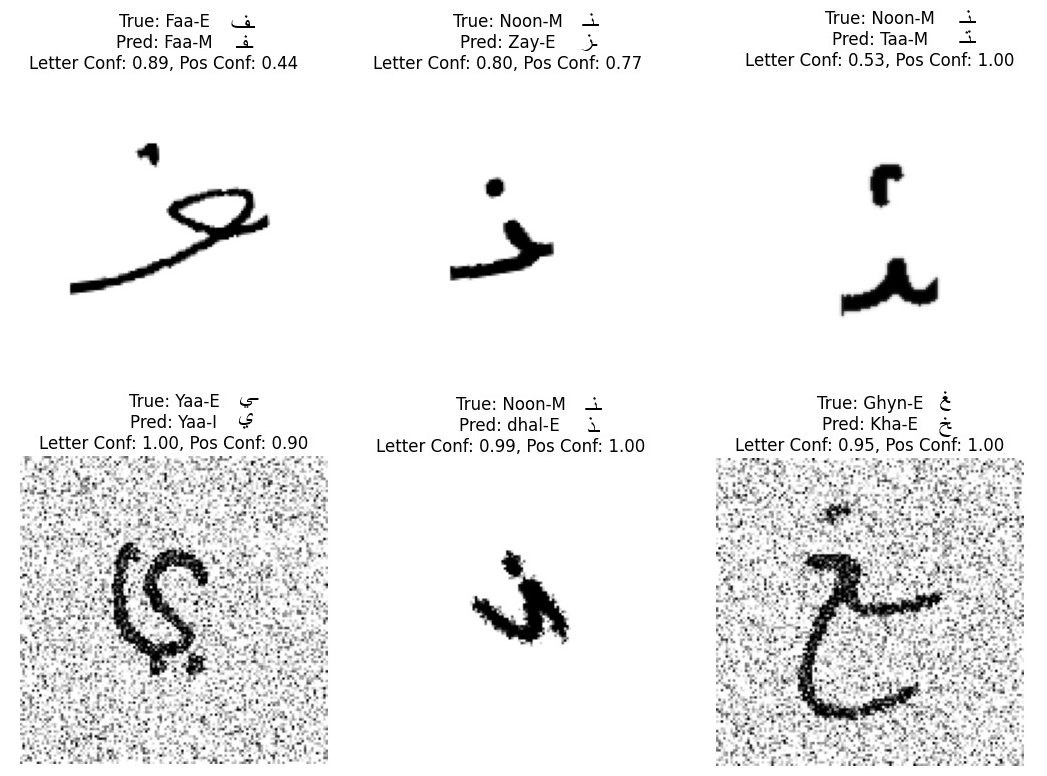}
    \caption{Examples of Handwriting-Induced Ambiguity in Arabic Script}
    \label{fig:Example_}
\end{figure}

A key observation is the considerable variability in how certain letters are written, sometimes making them visually ambiguous even for human observers. As illustrated in Figure~\ref{fig:Example_}, natural handwriting variations—such as differences in stroke curvature, thickness, and spacing—can create challenges that extend beyond machine recognition. The model's ability to achieve high accuracy despite these complexities highlights the effectiveness of the hybrid approach.

This work contributes to the advancement of OCR systems for Arabic script, providing a scalable solution with potential real-world applications. Future research could explore extending this framework to other cursive scripts or incorporating stroke-level information to enhance recognition accuracy. By addressing key challenges in handwritten Arabic recognition, this study represents a step forward in developing more robust and adaptable OCR technologies capable of handling diverse handwriting styles with greater reliability.

\bibliographystyle{IEEEtran}
\input{paper.bbl}

\end{document}